\documentclass[review]{elsarticle}
\usepackage{hyperref}

\usepackage{graphics}
\usepackage{graphicx}
\newsavebox\CBox

\usepackage{amsmath,amssymb,amsfonts, commath}
\usepackage{tabularx}
\usepackage{subcaption} 
\usepackage{booktabs}
\usepackage{siunitx}
\usepackage[capitalise]{cleveref}
\crefformat{appendix}{#2#1#3}
\usepackage[symbol]{footmisc}
\usepackage{diagbox}
\usepackage{xspace}
\usepackage{longtable}
\usepackage{float}

\usepackage{pifont}
\newcommand{\cmark}{\ding{51}}%
\newcommand{\xmark}{\ding{55}}%

\DeclareMathOperator*{\E}{\mathbb{E}}

\newcommand{\ExNLLShort}{$\E_{\theta}\mathrm{[NLL]}$\xspace}

\newcommand{\Infbae}{BAE-\scalebox{1.25}{$\infty$}\,}%

\usepackage[figurename=Fig.]{caption}

\journal{-}

\begin{document}

\begin{frontmatter}

\title{Do autoencoders need a bottleneck for anomaly detection?}

\author{Bang Xiang Yong, Alexandra Brintrup}
\address{Institute for Manufacturing, University of Cambridge, UK}

\begin{abstract}

A common belief in designing deep autoencoders (AEs), a type of unsupervised neural network, is that a bottleneck is required to prevent learning the identity function. Learning the identity function renders the AEs useless for anomaly detection. In this work, we challenge this limiting belief and investigate the value of non-bottlenecked AEs.

The bottleneck can be removed in two ways: (1) overparameterising the latent layer, and (2) introducing skip connections. However, limited works have reported on the use of one of the ways. For the first time, we carry out extensive experiments covering various combinations of bottleneck removal schemes, types of AEs and datasets. In addition, we propose the infinitely-wide AEs as an extreme example of non-bottlenecked AEs. 

Their improvement over the baseline implies learning the identity function is not trivial as previously assumed. Moreover, we find that non-bottlenecked architectures (highest AUROC=0.857) can outperform their bottlenecked counterparts (highest AUROC=0.696) on the popular task of CIFAR (inliers) vs SVHN (anomalies), among other tasks, shedding light on the potential of developing non-bottlenecked AEs for improving anomaly detection.







\end{abstract}

\begin{keyword}
\texttt{autoencoders, anomaly detection, bottleneck, unsupervised neural network}
\end{keyword}

\end{frontmatter}

\section{Introduction}

Numerous works have demonstrated the successful use of autoencoders (AEs), a type of unsupervised neural network (NN), for anomaly detection \cite{pang2021}. AEs are optimised to reconstruct a set of training data with minimal error. When given anomalous data which have high dissimilarity from the training data, the AEs reconstruct them with high error. Therefore, the reconstruction error is a measure of data anomalousness; by placing a threshold, we can effectively classify data points as inliers or anomalies.

Extant works claim that AEs will trivially learn the identity function when no constraints are placed \cite{ruff2018, ng2011sparse}. If this were to occur, AEs will perfectly reconstruct any inputs (regardless whether it is anomalous or not), and hence the reconstruction loss will be low for all inputs, leading to unreliable anomaly detection. To prevent this, it is common to impose a bottleneck in the architecture, resulting in an undercomplete architecture: the output of the encoder has much lower dimensions than the input. However, most works describe the need for a bottleneck analogically and report only the empirical performance of bottlenecked AEs, without comparing them against non-bottlenecked AEs \cite{chen2018autoencoder, chow2020, kim2020, mujeeb2019, ruff2018}. 

\textbf{Why should we care about non-bottlenecked AEs?} By limiting to bottlenecked architectures, we miss the potential of achieving better performance with non-bottlenecked AEs. Therefore, in this work, we study the use of non-bottlenecked AEs for anomaly detection. We investigate combinations of ways for removing the bottleneck, including (1) expanding the latent dimensions and (2) introducing skip connections. Furthermore, we propose the infinitely-wide AEs as an extreme example. Extensive experiments demonstrate the empirical success of non-bottlenecked AEs in detecting anomalies over the baseline and the bottlenecked AEs, indicating the non-bottlenecked AEs have failed to learn the identity function, contrary to conventional belief. 

\begin{figure}[hbtp]
\centerline{\includegraphics[width=\columnwidth]{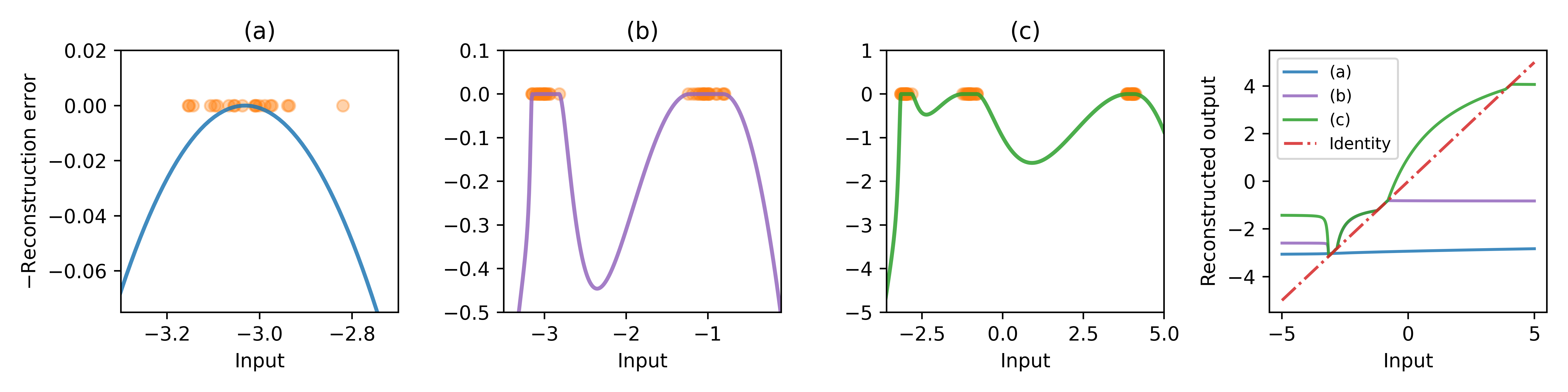}}
\caption{
(a-c) Negative reconstruction errors (log-likelihood) from BAEs with five layers of infinitely many parameters on 1D toy datasets, resembling reasonable density estimation. Orange dots represent the training data points. (d) The reconstructed outputs (last panel) clearly differ from the identity function. All layers use GELU \cite{hendrycks2016gaussian} as activation functions, except the last, which uses the sigmoid function; min-max scaler \cite{scikit-learn} is used.  
}
\label{fig:inf-bae-1d}
\end{figure}

We suggest that rethinking about AEs is needed. In this effort, we adopt the probabilistic formulation of Bayesian autoencoders (BAEs), viewing them as regularised density estimators that benefit from having higher expressivity allowed by non-bottlenecked architectures (see \cref{fig:inf-bae-1d} for an example). The Bayesian framework also provides a sound foundation for theoretical analysis of these architectures in future work.

This paper is organised as follows: \cref{sec:methods} formulates AEs from a Bayesian perspective and describes ways to remove the bottleneck. Our experimental setup is described in \cref{sec:exp-setup} followed by results and discussion in \cref{sec:res-disc}. We relate to previous works in \cref{sec:related} and state our limitations in \cref{sec:limits}. We close with a summary and future directions in \cref{sec:conclusion}.

\section{Methods} \label{sec:methods}

\subsection{Bayesian autoencoders}
Suppose we have a set of data $X^{train} = {\{\textbf{x}_1,\textbf{x}_2,\textbf{x}_3,...\textbf{x}_N\}}$, $\textbf{x}_{i} \in \rm I\!R^{D}$. An AE is an NN parameterised by $\theta$, and consists of two parts: an encoder for mapping input data $\textbf{x}$ to a latent embedding, $\textbf{z}=f_\text{encoder}(\textbf{x})$, and a decoder $f_\text{decoder}$ for mapping the latent embedding to a reconstructed signal of the input $\hat{\textbf{x}}$ (i.e. $\hat{\textbf{x}} = {f_{\theta}}(\textbf{x}) = f_\text{decoder}(f_\text{encoder}(\textbf{x}))$) \cite{goodfellow2016deep}. 

Bayes' rule can be applied to the parameters of the AE to create a BAE,
\begin{equation}\label{eq_posterior}
    p(\theta|X^{train}) = \frac{p(X^{train}|\theta)\, p(\theta)}{p(X^{train})} \\ ,
\end{equation}
where $p(X^{train}|\theta)$ is the likelihood and $p(\theta)$ is the prior distribution of the AE parameters. The log-likelihood for a diagonal Gaussian distribution is,
\begin{equation} \label{eq_gaussian_loss}
\log{p(\textbf{x}|\theta)} = -(\frac{1}{D} \sum^{D}_{i=1}{\frac{1}{2\sigma_i^2}}(x_i-\hat{x_i})^2 + \frac{1}{2}\log{\sigma_i^2})
\end{equation} 
where $\sigma_i^2$ is the variance of the Gaussian distribution. For simplicity, we use an isotropic Gaussian likelihood with $\sigma_i^2=1$ in this study, since the negative log-likelihood (NLL) is proportional to the mean-squared error (MSE) function. We employ an isotropic Gaussian prior distribution, effectively leading to $L_2$ regularisation.

Since Equation~\ref{eq_posterior} is analytically intractable for a deep NN, various approximate methods have been developed such as Stochastic Gradient Markov Chain Monte Carlo (SGHMC) \cite{chen2014stochastic}, Monte Carlo Dropout (MCD) \cite{gal2016dropout}, Bayes by Backprop (BBB) \cite{blundell2015weight}, and anchored ensembling \cite{pearce2020uncertainty} to sample from the posterior distribution. In contrast, a deterministic AE has its parameters estimated using maximum likelihood estimation (MLE) or maximum a posteriori (MAP) when regularisation is introduced. The variational autoencoder (VAE) \cite{kingma2013auto} and BAE are AEs formulated differently within a probabilistic framework: in the VAE, only the latent embedding is stochastic while the $f_\text{encoder}$ and $f_\text{decoder}$ are deterministic and the model is trained using variational inference; on the other hand, the BAE, as an unsupervised Bayesian neural network (BNN), has distributions over all parameters of $f_\text{encoder}$ and $f_\text{decoder}$. In short, the training phase of BAE entails using one of the sampling methods to obtain a set of approximate posterior samples $\{\hat{\theta}_m\}^M_{m=1}$. 

Then, during the prediction phase, we use the posterior samples to compute $M$ estimates of the NLL. The predictive density of a new data point $\mathbf{x}^*$ can be approximated as the mean of the posterior NLL estimates,
\begin{equation}
p(\textbf{x}^*| X^{train}) = \E{}_{\theta}[-\log{p(\textbf{x}^*|\theta)\,p(\theta|X^{train})}] \approx 
-\frac{1}{M}\sum^M_{m=1}p(\textbf{x}^*|\hat{\theta}_m)
\end{equation}
For convenience, we denote $p(\textbf{x}^*| X^{train})$ as \ExNLLShort. The Bayesian formulation allows us to view AEs as regularised probability density estimators: they model the training data distribution, assigning lower density scores to data which have higher dissimilarity from the training data. 

\subsection{How to remove the bottleneck?}
\begin{figure}[hbtp]
    \begin{subfigure}[b]{0.5\textwidth}
    \includegraphics[width=\textwidth]{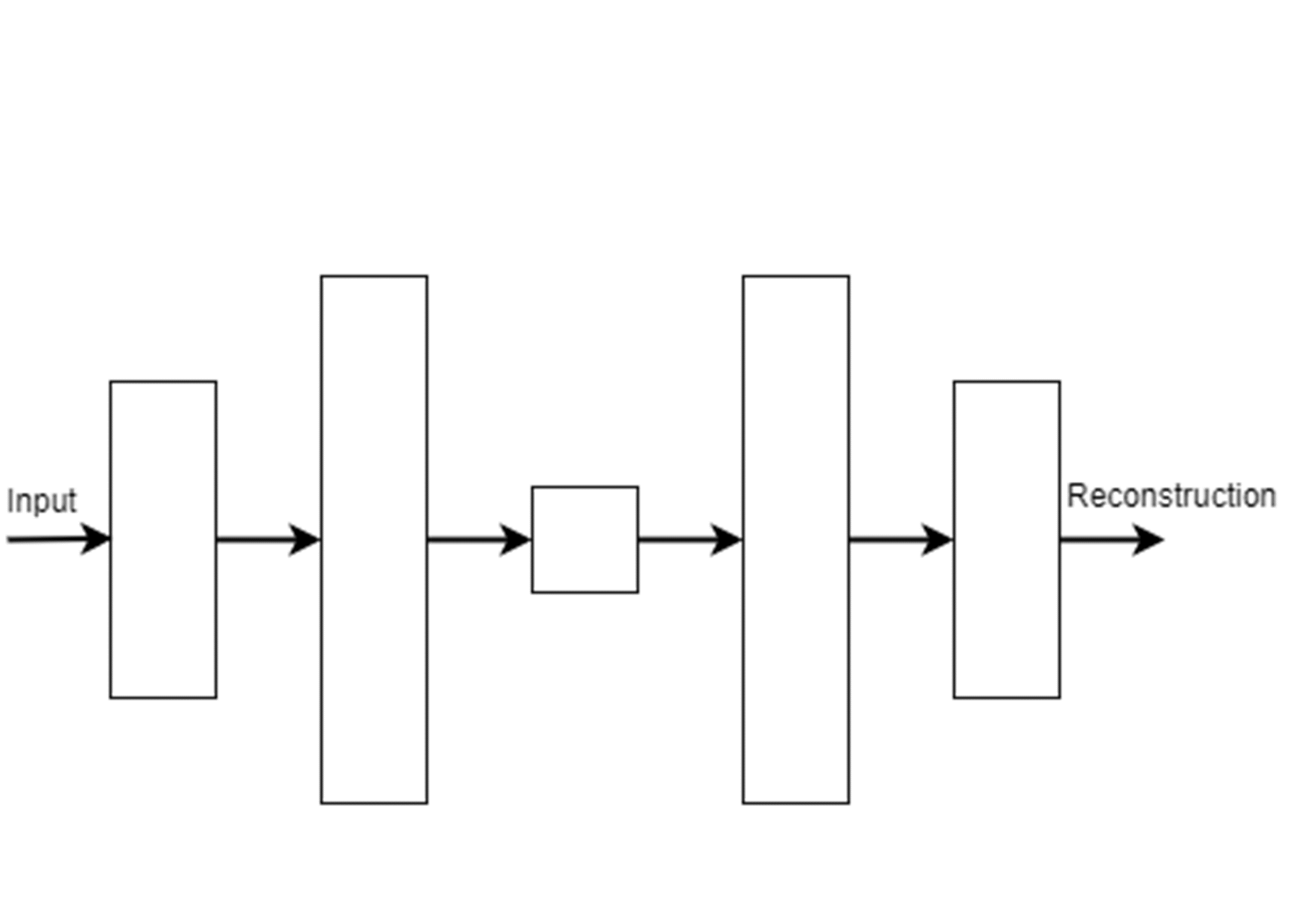}
    \centering
    \caption{Undercomplete, no skip connections}
    \label{1}
    \end{subfigure}
    \begin{subfigure}[b]{0.5\textwidth}
    \includegraphics[width=\textwidth]{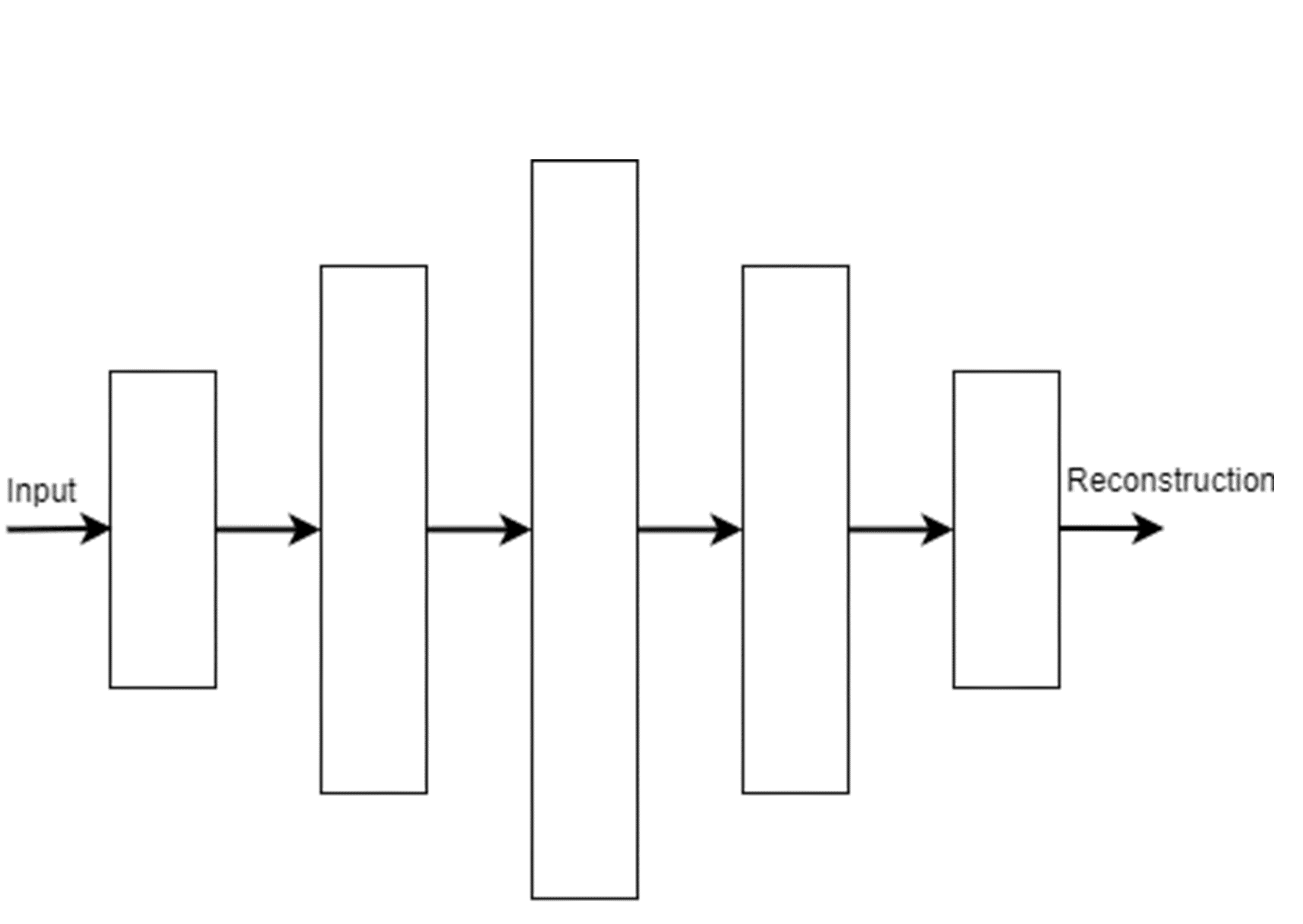}
    \centering
    \caption{Overcomplete, no skip connections}
    \label{2}
    \end{subfigure}
        \begin{subfigure}[b]{0.5\textwidth}
    \includegraphics[width=\textwidth]{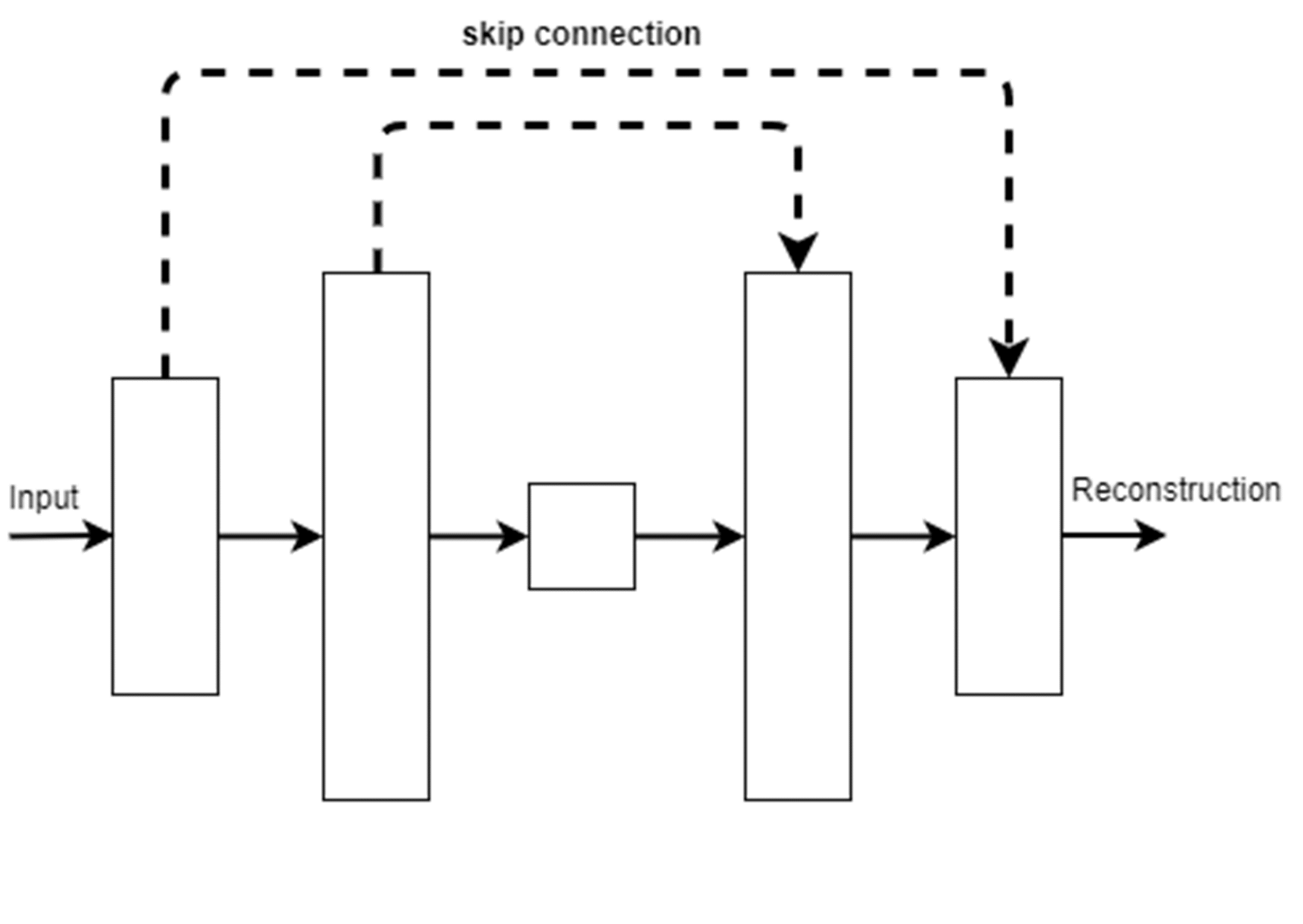}
    \centering
   \caption{Undercomplete +skip connections}
    \label{3}
    \end{subfigure}
    \begin{subfigure}[b]{0.5\textwidth}
    \includegraphics[width=\textwidth]{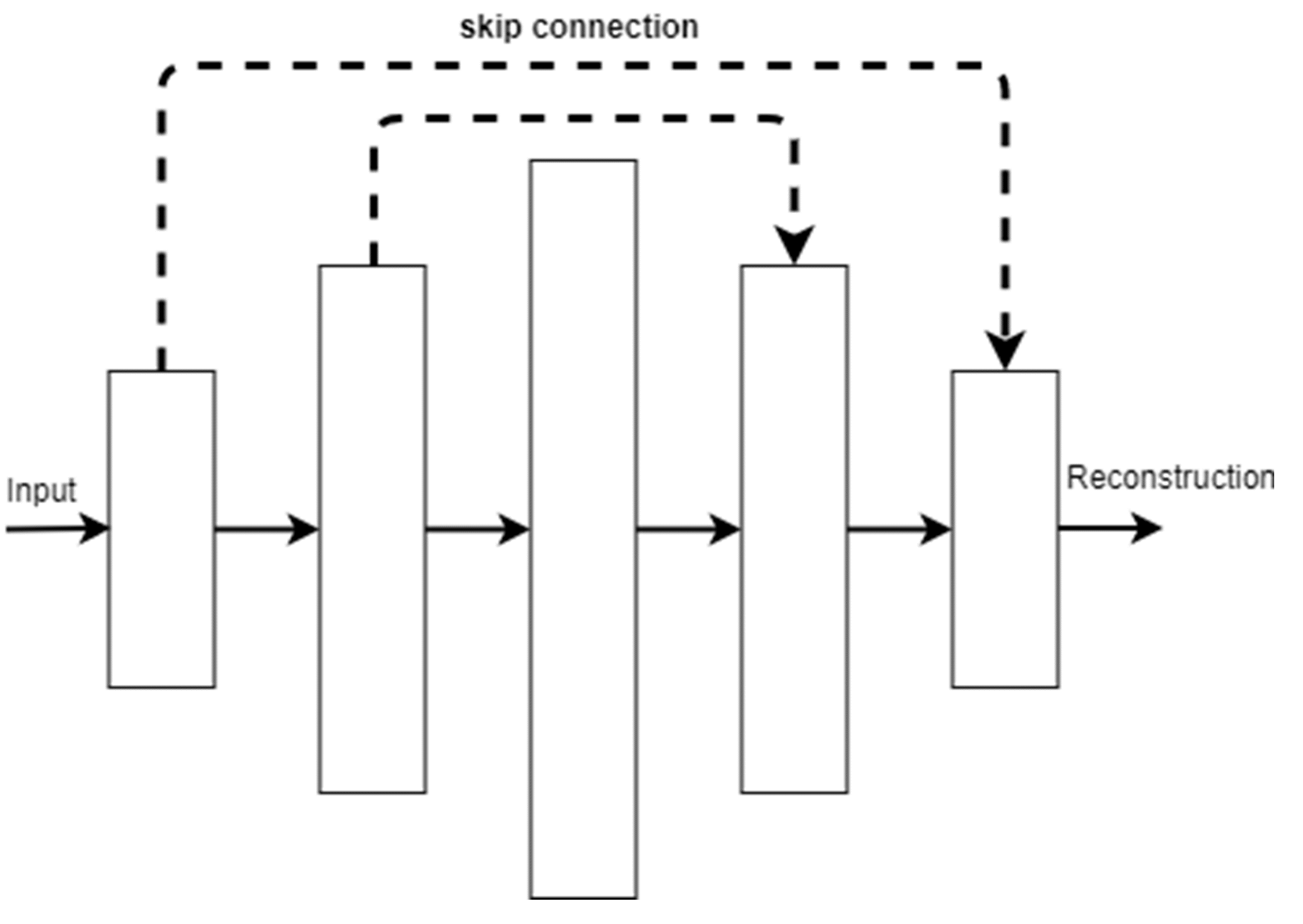}
    \centering
    \caption{Overcomplete +skip connections}
    \label{4}
    \end{subfigure}
    \caption{Architectures of bottlenecked (type A) and non-bottlenecked (type B, C and D) AEs. Each block represents an NN layer. Its width indicates the relative number of parameters.}
    \label{fig:arc-bottle}
\end{figure}

The identity function is successfully learnt when $f_{\theta}(\textbf{x}) = \textbf{x}$ holds true for all \textbf{x} and therefore the reconstruction loss or NLL is always 0, rendering it useless for distinguishing anomalies from inliers. In an effort to mitigate this, a bottleneck is implemented at the latent layer (encoder's final layer) by having the latent dimensions smaller than the input dimensions, $\text{dim}(\textbf{z})<\text{dim}(\textbf{x})$, and there is no way for any output of the intermediate layers to bypass the bottleneck layer. It is straightforward to eliminate the bottleneck by doing the opposite: (1) simply expand the size of the latent dimensions to $\text{dim}(\textbf{z})\ge{\text{dim}(\textbf{x})}$, also known as an overcomplete architecture, and/or (2) introduce long-range skip connections from the encoder to the decoder akin to a U-Net architecture \cite{unet2015}, thereby allowing each layer's data flow to bypass the bottleneck; for clarity, see \cref{table:arc-bottle} and \cref{fig:arc-bottle}. 

\begin{table}[hbtp]
\caption{Categorising architectures into with or without a bottleneck depends on the latent dimensions and the presence of skip connections.}
\centering
\resizebox{0.80\textwidth}{!}{
    \begin{tabular}{ccc}
    \hline
    Architecture type & Latent dimensions & Skip connections \\ \hline
    \textbf{Bottlenecked} &  &  \\
    A & Undercomplete & \xmark \\ \hline
    \textbf{Non-bottlenecked} &  &  \\
    B & Undercomplete & \cmark \\
    C & Overcomplete & \xmark \\
    D & Overcomplete & \cmark \\ \hline
    \end{tabular} 
    }
    \label{table:arc-bottle}
\end{table}

\textbf{Why skip connections?} Skip connections allow a better flow of information in NNs with many layers, leading to a smoother loss landscape \cite{li2017visualizing} and easier optimisation, without additional computational complexity \cite{he2016deep}. Recent works \cite{baur2020bayesian, collin2021, kim2021} have reported that AEs with skip connections outperform those without on image anomaly detection. In preventing the skip-AEs from learning the identity function, Collin et al. \cite{collin2021} and Baur et al. \cite{baur2020bayesian} have implemented a denoising scheme and a dropout mechanism, respectively. Notably, Baur et al. \cite{baur2020bayesian} have reported that random weight initialisation alone is sufficient to prevent learning the identity function, rendering the dropout redundant. 

\textbf{Infinitely-wide BAE.} In the infinite-width limit, a fully-connected BNN is equivalent to a neural network Gaussian process (NNGP) \cite{neal1996priors}. The results have been extended to modern architectures such as convolutional NNs, recurrent NNs, and transformers \cite{novak2018bayesian, yang2019nngp, hron2020infinite} in recent years. We propose extending the NNGP to the AE to create an infinitely-wide BAE (\Infbae), which opposes the conventional bottleneck design. Viewing the BAE as a density estimator motivates this; it is not unconventional for density estimators to have infinite parameters as they benefit from higher expressivity to model an arbitrary distribution well \cite{chen2006probability, sriperumbudur2017density}. 
\begin{figure}[hbtp]
\centerline{\includegraphics[width=\columnwidth]{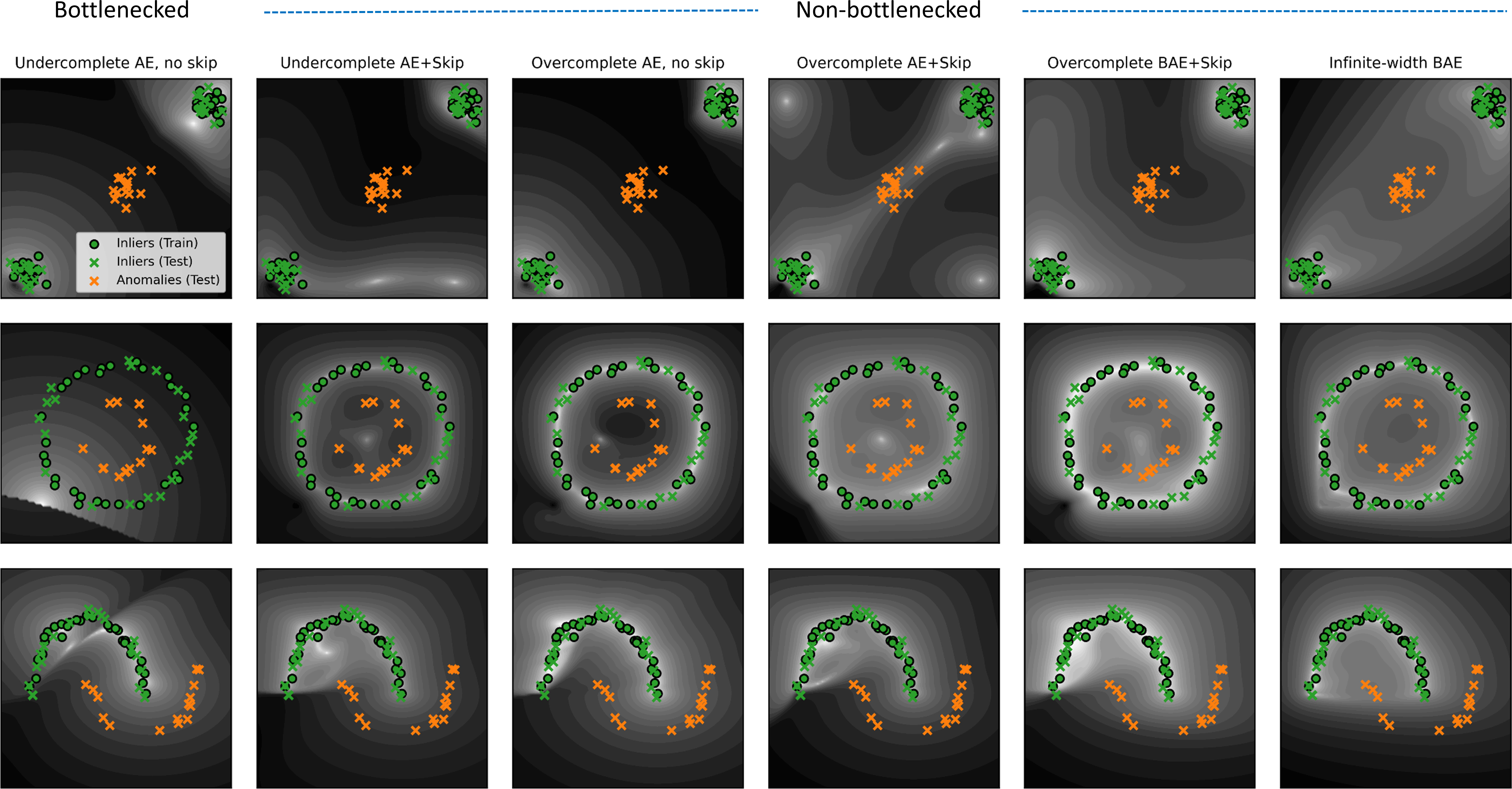}}
\caption{\ExNLLShort using deterministic AE and BAEs with bottlenecked and non-bottlenecked architectures. Brighter region has lower \ExNLLShort values and log-scale of contour is used to increase visibility. The encoder architecture has fully-connected layers with nodes of 2-50-50-50-dim(\textbf{z}) where dim(\textbf{z})=1 for undercomplete and dim(\textbf{z})=100 for overcomplete architectures. We use SELU activation \cite{klambauer2017self} for every layer and sigmoid activation for the final layer. The \Infbae has a similar number of layers with infinite parameters. The darker contours away from the training points show that the deterministic AE and BAE do not learn the identity function despite being overparameterised and having skip connections.}
\label{fig:toy-bottleneck-2d}
\end{figure}
There are two primary advantages of the NNGP: having a closed-form solution and modelling a BNN with infinitely many parameters. The first facilitates a theoretical understanding by linking to the well-studied GP model, and the second \textit{potentially} improves performance since deep NNs succeed over traditional ML models via increasing model parameters \cite{pang2021}. Nonetheless, empirically, infinite NNs do not always outperform finite NNs; reasons for their underperformance remain an active research topic \cite{aitchison2020bigger, lee2020finite}. Another drawback is their computational complexity of $\mathcal{O}(N^3)$, where $N$ is the number of training examples, reducing scalability to large datasets.

Surprisingly, when we examine the behaviours of AEs on 2D toy data sets  (\cref{fig:toy-bottleneck-2d}), we find that the identity function is not learnt despite using various types of non-bottlenecked AEs. Consequently, this observation on low-dimensional data implies it is more unlikely to learn the identity function on high-dimensional data due to higher degrees of freedom.

We suggest several reasons hindering AEs from the identity mapping: high degree of non-linearity in the AE and regularisation induced by mini-batching, the deep learning optimiser (e.g. Adam \cite{kingma2014adam}) and the prior over parameters. Since these are usually implicit in training the AE, no additional, explicit efforts are necessary (e.g. denoising or dropout mechanisms). 

\section{Experimental Setup} \label{sec:exp-setup}

The full code for reproducing results will be released upon the acceptance of this paper.

\subsection{Datasets}
Several publicly available datasets are included in our experiments. For image data, we use pairs of unrelated datasets (inliers vs anomalies) commonly used in previous works: FashionMNIST \cite{xiao2017fashion} vs MNIST \cite{deng2012mnist}, and CIFAR \cite{krizhevsky2009learning} vs SVHN \cite{netzer2011}. For tabular data, we use eight datasets from the Outlier Detection Datasets (ODDS) collection \cite{rayana2016}: Cardio, Lympho, Optdigits, Ionosphere, Pendigits, Thyroid, Vowels and Pima. For sensors data, we use the ZeMA \cite{tizian2018} and STRATH \cite{tachtatzis2019} datasets gathered from industrial environments.

In the ZeMA dataset, the tasks are to detect deterioration in the subsystems of a hydraulic test rig (see \cref{table:des_zema_strath} for the sensor-subsystem pairs). We consider the data from the healthiest state of the subsystems as inliers and the rest as anomalies. In the STRATH dataset, different sensors are used in each task to detect defective parts manufactured from a radial forging process consisting of heating and forging phases. Geometric measurements of each forged part are available as target quality indicators. To label the anomalies, we focus our analysis on the \textit{38 diameter@200} by applying the Tukey's fences method \cite{tukey1977exploratory} on the absolute difference between the measured and nominal dimensions. 

\subsection{Preprocessing}
For image data, we use the default split of train-test sets and rescale pixel values to [0,1]. For the ODDS, ZeMA and STRATH datasets, we split the inliers into train-test sets of 70:30 ratio with random shuffling, and include all anomalies in the test set. We apply min-max scaling \cite{scikit-learn} with care to prevent train-test bias by fitting the scaler to the train set instead of the entire inlier set. For ZeMA, we downsample the pressure sensor to 1Hz and use the temperature data as provided; for STRATH, we downsample the data by tenfold and segment only the forging phase.  

\begin{table}[]
\centering
\caption{Sensors used for each task in ZeMA and STRATH datasets.}

\resizebox{0.80\textwidth}{!}{
        \begin{tabular}{@{}clll@{}}
        \toprule
        \multicolumn{1}{l}{} & \multicolumn{2}{c}{ZeMA} & \multicolumn{1}{c}{STRATH} \\
        \multicolumn{1}{l}{Tasks} & \multicolumn{1}{c}{Target subsystem} & \multicolumn{1}{c}{Sensors} & \multicolumn{1}{c}{Sensors} \\ \midrule
        (i) & Cooler & Temperature (TS4) & Position (L-ACTpos) \\
        (ii) & Valve & Temperature (TS4) & Speed (A-ACTspd) \\
        (iii) & Pump & Pressure      (PS6) & Servo (Feedback-SPA) \\
        (iv) & Accumulator & Temperature (TS4) & Sensors from tasks (i-iii) \\ \bottomrule
        \end{tabular}
}
\label{table:des_zema_strath}
\end{table}

\subsection{Models}
We train variants of AEs: deterministic AE, VAE, and BAEs with several inference methods: MCD, BBB and anchored ensembling. All models use isotropic Gaussian priors over weights. The number of posterior samples is set to $M=100$ for the VAE, BAE-MCD and BAE-BBB, while $M=10$ for the BAE-Ensemble. We set a fixed learning rate of 0.001 for STRATH; the learning rates for the other datasets are searched with a learning rate finder, employing a cyclic learning rate \cite{smith2017cyclical}. The Adam optimiser is used and the training epochs for FashionMNIST, CIFAR, ODDS, ZeMA and STRATH are 20, 20, 300, 100, 100, respectively. The weight decay is set to $1\times{10^{-11}}$ for FashionMNIST and CIFAR, and $1\times{10^{-10}}$ for ODDS, ZeMA and STRATH.

\begin{table}[H] 
\centering
\caption{Encoder architecture of finite-width AEs. The decoder is a reflection of the encoder, in which the Conv1D and Conv2D layers are replaced by Conv1D- and Conv2D-Transpose layers. The leaky ReLu \cite{maas2013rectifier} is used as the activation function with a slope of 0.01 while the sigmoid function is used at the decoder's final layer.}

    \begin{subtable}[h]{\textwidth}
        \centering
        \vskip 0.05in
        \scriptsize \caption{FashionMNIST and CIFAR}
            \resizebox{0.60\textwidth}{!}{
                \begin{tabular}{@{}cccc@{}}
                \toprule
                Layer       & Output channels/nodes & Kernel & Strides \\ \midrule
                Conv2D & 10      & 2 x 2  & 2 x 2   \\
                Conv2D & 32      & 2 x 2  & 1 x 1   \\
                Reshape     & -             & -      & -       \\
                Dense       & 100       & -      & -      \\
                Dense       & Latent dimensions       & -      & -      \\ \bottomrule
                \end{tabular}
            }
    \end{subtable}
    \hfill
    \begin{subtable}[h]{\textwidth}
        \centering
        \vskip 0.1in
        \scriptsize \caption{ODDS}
            \resizebox{0.60\textwidth}{!}{
                    \begin{tabular}{@{}cccc@{}}
                    \toprule
                    Layer       & Output channels/nodes & Kernel & Strides \\ \midrule
                    Linear       & Input dimensions $\times 4$      & -      & -      \\ 
                    Linear       & Input dimensions $\times 4$       & -      & -      \\ 
                    Linear       & Latent dimensions       & -      & -      \\ \bottomrule
                    \end{tabular}
            }
    \end{subtable}
    \hfill
    \begin{subtable}[h]{\textwidth}
        \centering
        \vskip 0.1in
        \scriptsize \caption{ZeMA and STRATH}
            \resizebox{0.60\textwidth}{!}{
                \begin{tabular}{@{}cccc@{}}
                \toprule
                Layer       & Output channels/nodes & Kernel & Strides \\ \midrule
                Conv1D & 10      & 8  & 2   \\
                Conv1D & 20      & 2  & 2   \\
                Reshape     & -              & -      & -       \\
                Linear       & 1000       & -      & -      \\ 
                Linear       & Latent dimensions       & -      & -      \\ \bottomrule
                \end{tabular}
            }
    \end{subtable}
     \label{table:all-architecture}
\end{table}


The architectures of non-finite AEs are described in \cref{table:all-architecture}. All architectures, except for ZeMA, apply layer normalisation \cite{ba2016layer} before the activation function. Leaky ReLu \cite{maas2013rectifier} is used as the intermediate layers' activation function while the sigmoid function is used at the final output layer. The bias terms are turned off for each layer. The size of latent dimensions is set to the flattened input dimensions multiplied by factors of $\times{\frac{1}{10}}, \times{\frac{1}{2}}, \times{1}$, including $\times{2}$ (for FashionMNIST and CIFAR) and $\times{10}$ (for ODDS, ZeMA and STRATH). For the \Infbae, we implement the NNGP with seven infinitely-wide dense layers (including the encoder and decoder) using the Neural Tangent Kernel library \cite{neuraltangents2020} for all datasets. During testing, we evaluate the area under the receiver-operating characteristic curve (AUROC) \cite{melo2013} scores of the \ExNLLShort on the test set consisting of inliers and anomalies.

\section{Results and discussion} \label{sec:res-disc}

In \cref{table:btneck-res}, most non-bottlenecked models (type B, C and D) beat the baseline with mean AUROC $\ge{0.8}$. This observation indicates the identity function has not been learnt despite being overparameterised and having skip connections. Also, their positive average treatment effect (ATE) improves over the bottlenecked models (type A) with type D models showing the highest ATE. In addition, the best mean AUROC scores on most datasets have been achieved by the non-bottlenecked models, except on FashionMNIST vs MNIST, which has the bottlenecked model (BAE-BBB, type A) marginally beating the second-best model (BAE-BBB, type C) by 0.1 AUROC.

Focusing on CIFAR vs SVHN, our results provide new insights into previous works which reported poor performances \cite{nalisnick2018do, choi2018waic}. Notably, the best non-bottlenecked model (BAE-Ensemble, type B, AUROC=0.849) and the \Infbae (AUROC=0.771) outperform the best bottlenecked model (BAE-Ensemble, type A, AUROC=0.696). These results imply the poor performance could be fixed if previous works were to consider non-bottlenecked architectures.

Switching from a deterministic AE to a BAE improves performance as the best performing BAEs achieve the highest AUROC scores on all datasets. The performance gain is attributed to Bayesian model averaging \cite{hinne2020bma}, which accounts for uncertainty in model parameters. The best BAEs also outperform the VAEs, evidencing the advantage of addressing the uncertainty over parameters of the entire model instead of considering only the latent layer.

\begin{figure}[hbtp]
\centerline{\includegraphics[width=\columnwidth]{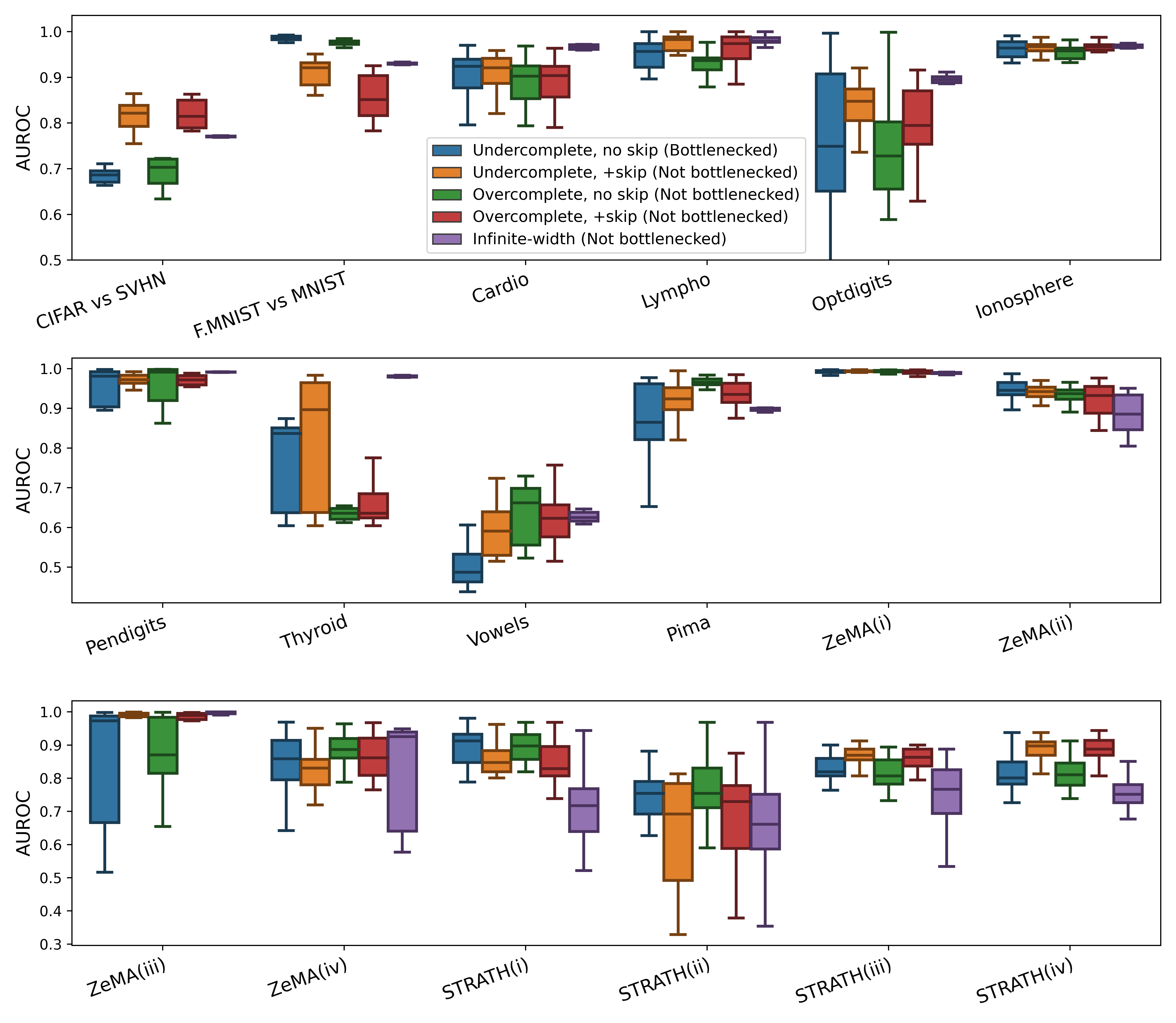}}
\caption{AUROC scores of bottlenecked and non-bottlenecked BAEs on various datasets. Results are shown for the finite-width BAE-Ensemble and the infinite-width BAE.}
\label{fig:boxplots-bottleneck}
\end{figure}

Although the \Infbae does not score the highest AUROC, on a positive note, there are specific tasks on which the \Infbae outperforms other models with the highest median AUROC (e.g. see \cref{fig:boxplots-bottleneck} on Cardio, Thyroid, Pendigits and ZeMA(iii)) and with low variability in performance. However, the gain over the finite-width BAEs is not demonstrated on some tasks of ZEMA and STRATH.

\begin{table}[H]
\caption{Mean\,$\pm$\,standard error AUROC scores for deterministic AEs, VAEs and BAEs with bottlenecked and non-bottlenecked architectures. Model with highest mean AUROC is bolded for each dataset.}
\centering
\resizebox{.78\textwidth}{!}{
\begin{tabular}{@{}lcccc@{}}
                \toprule
                 & ---Bottlenecked--- & \multicolumn{3}{c}{------------------Not bottlenecked------------------} \\ 
                 \multicolumn{1}{c}{Model} & \begin{tabular}[c]{@{}c@{}}A\\ Undercomplete\\ No skip\end{tabular} & \begin{tabular}[c]{@{}c@{}}B\\ Undercomplete\\ + skip\end{tabular} & \begin{tabular}[c]{@{}c@{}}C\\ Overcomplete\\ No skip\end{tabular} & \begin{tabular}[c]{@{}c@{}}D\\ Overcomplete\\ + skip\end{tabular} \\ \midrule

\multicolumn{5}{l}{\textbf{CIFAR vs SVHN}, four runs} \\
\,\quad Deterministic AE & 0.686\,$\pm$\,0.006 & 0.816\,$\pm$\,0.013 & 0.692\,$\pm$\,0.012 & 0.820\,$\pm$\,0.011 \\
\,\quad VAE & 0.398\,$\pm$\,0.006 & 0.822\,$\pm$\,0.012 & 0.428\,$\pm$\,0.004 & 0.840\,$\pm$\,0.008 \\
\,\quad BAE-MCD & 0.613\,$\pm$\,0.012 & 0.809\,$\pm$\,0.014 & 0.626\,$\pm$\,0.012 & 0.834\,$\pm$\,0.004 \\
\,\quad BAE-BBB & 0.672\,$\pm$\,0.006 & 0.831\,$\pm$\,0.016 & 0.639\,$\pm$\,0.007 & 0.826\,$\pm$\,0.006 \\
\,\quad BAE-Ensemble & 0.696\,$\pm$\,0.009 & \textbf{0.849\,$\pm$\,0.005} & 0.699\,$\pm$\,0.008 & 0.838\,$\pm$\,0.005 \\
\,\quad BAE-\scalebox{1.25}{$\infty$} & - & - & 0.771\,$\pm$\,0.001 & - \\
\multicolumn{5}{l}{\textbf{FashionMNIST vs MNIST}, four runs} \\
\,\quad Deterministic AE & 0.986\,$\pm$\,0.002 & 0.912\,$\pm$\,0.012 & 0.976\,$\pm$\,0.002 & 0.858\,$\pm$\,0.019 \\
\,\quad VAE & 0.793\,$\pm$\,0.02 & 0.895\,$\pm$\,0.014 & 0.913\,$\pm$\,0.008 & 0.884\,$\pm$\,0.021 \\
\,\quad BAE-MCD & 0.983\,$\pm$\,0.001 & 0.712\,$\pm$\,0.033 & 0.984\,$\pm$\,0.001 & 0.783\,$\pm$\,0.048 \\
\,\quad BAE-BBB & \textbf{0.992\,$\pm$\,0.001} & 0.848\,$\pm$\,0.027 & 0.99\,$\pm$\,0.002 & 0.901\,$\pm$\,0.009 \\
\,\quad BAE-Ensemble & 0.985\,$\pm$\,0.003 & 0.920\,$\pm$\,0.006 & 0.977\,$\pm$\,0.001 & 0.928\,$\pm$\,0.009 \\
\,\quad BAE-\scalebox{1.25}{$\infty$} & - & - & 0.930\,$\pm$\,0.002 & - \\
\multicolumn{5}{l}{\textbf{ODDS}, eight datasets, ten runs} \\
\,\quad Deterministic AE & 0.819\,$\pm$\,0.014 & 0.872\,$\pm$\,0.011 & 0.833\,$\pm$\,0.013 & 0.848\,$\pm$\,0.012 \\
\,\quad VAE & 0.795\,$\pm$\,0.014 & 0.876\,$\pm$\,0.008 & 0.819\,$\pm$\,0.013 & 0.883\,$\pm$\,0.008 \\
\,\quad BAE-MCD & 0.829\,$\pm$\,0.015 & 0.891\,$\pm$\,0.010 & 0.819\,$\pm$\,0.013 & 0.856\,$\pm$\,0.011 \\
\,\quad BAE-BBB & 0.892\,$\pm$\,0.011 & 0.913\,$\pm$\,0.007 & \textbf{0.919\,$\pm$\,0.007} & 0.917\,$\pm$\,0.007 \\
\,\quad BAE-Ensemble & 0.857\,$\pm$\,0.012 & 0.886\,$\pm$\,0.010 & 0.850\,$\pm$\,0.011 & 0.86\,$\pm$\,0.012 \\
\,\quad BAE-\scalebox{1.25}{$\infty$} & - & - & 0.913\,$\pm$\,0.013 & - \\
\multicolumn{5}{l}{\textbf{ZeMA}, four tasks, ten runs} \\
\,\quad Deterministic AE & 0.879\,$\pm$\,0.024 & 0.930\,$\pm$\,0.01 & 0.911\,$\pm$\,0.014 & 0.908\,$\pm$\,0.019 \\
\,\quad VAE & 0.879\,$\pm$\,0.015 & 0.917\,$\pm$\,0.016 & 0.876\,$\pm$\,0.019 & 0.922\,$\pm$\,0.016 \\
\,\quad BAE-MCD & 0.892\,$\pm$\,0.017 & 0.920\,$\pm$\,0.015 & 0.862\,$\pm$\,0.023 & 0.936\,$\pm$\,0.008 \\
\,\quad BAE-BBB & 0.888\,$\pm$\,0.019 & 0.939\,$\pm$\,0.007 & 0.895\,$\pm$\,0.019 & 0.931\,$\pm$\,0.009 \\
\,\quad BAE-Ensemble & 0.938\,$\pm$\,0.007 & 0.961\,$\pm$\,0.005 & 0.928\,$\pm$\,0.007 & \textbf{0.963\,$\pm$\,0.004} \\
\,\quad BAE-\scalebox{1.25}{$\infty$} & - & - & 0.926\,$\pm$\,0.01 & - \\
\multicolumn{5}{l}{\textbf{STRATH}, four tasks, ten runs} \\
\,\quad Deterministic AE & 0.817\,$\pm$\,0.010 & 0.811\,$\pm$\,0.015 & 0.819\,$\pm$\,0.01 & 0.819\,$\pm$\,0.012 \\
\,\quad VAE & 0.825\,$\pm$\,0.009 & 0.831\,$\pm$\,0.012 & 0.831\,$\pm$\,0.008 & 0.824\,$\pm$\,0.014 \\
\,\quad BAE-MCD & 0.838\,$\pm$\,0.008 & 0.808\,$\pm$\,0.016 & 0.838\,$\pm$\,0.007 & 0.822\,$\pm$\,0.014 \\
\,\quad BAE-BBB & 0.847\,$\pm$\,0.008 & 0.834\,$\pm$\,0.007 & 0.848\,$\pm$\,0.008 & 0.833\,$\pm$\,0.008 \\
\,\quad BAE-Ensemble & 0.839\,$\pm$\,0.008 & 0.850\,$\pm$\,0.007 & 0.835\,$\pm$\,0.007 & \textbf{0.855\,$\pm$\,0.006} \\
\,\quad BAE-\scalebox{1.25}{$\infty$} & - & - & 0.717\,$\pm$\,0.009 & - \\ \hline
\quad Mean & 0.825\,$\pm$\,0.01 & 0.866\,$\pm$\,0.012 & 0.835\,$\pm$\,0.009 & \textbf{0.867\,$\pm$\,0.012} \\
\quad ATE & - & 0.041 & 0.010 & \textbf{0.042} \\ \hline
\end{tabular}
}
\label{table:btneck-res}
\end{table}

\section{Related work} \label{sec:related}
Several works have investigated the use of skip connections in AEs for tasks such as image denoising \cite{zhao2018,mao2016} and audio separation \cite{liu2018}. Our work differs from current works on skip-AEs for anomaly detection \cite{kim2021, baur2020bayesian, collin2021}: we have investigated a wider range of non-bottlenecked AEs, in which skip-AEs are only one type, and experimented with more datasets. 


Snoek et al. \cite{snoek2012nonparametric} has proposed the autoencoder with an infinitely-wide decoder while keeping its encoder finite, and demonstrates its effectiveness for supervised classification and learning latent representations. Nguyen et al. \cite{nguyen2021benefits} has theoretically studied infinitely-wide and shallow (two layers) AEs, providing insights into their behaviours. To the best of our knowledge, we are the first to propose a deep (seven layers) \Infbae with all layers being infinitely-wide, and provide empirical results on anomaly detection. 

Radhakrishnan et al. and Zhang et al. \cite{radhakrishnan2018memorization, zhang2019identity} have observed that overcomplete AEs exhibit \textit{memorisation}, a phenomenon where the AEs reconstruct the closest training examples instead of the inputs. We suggest that a possible link exists between memorisation and the success of detecting anomalies using overcomplete AEs: when given an anomalous input, the AEs reconstruct the closest training example of inliers. This leads to a more discriminating, larger reconstruction error with the anomalous input than if the input were to be an inlier. 

\section{Limitations} \label{sec:limits}
Our study has focused on unsupervised anomaly detection and implies nothing about other use cases (e.g. clustering and dimensionality reduction), for which a bottleneck is necessary. Our experiments have covered various data types, however, there may exist datasets where learning the identity function is trivial for the AE. While we lack theoretical proof that non-bottlenecked AEs never learn the identity function, the contrary is true; there is no proof, to the best of our knowledge, that they always learn the identity function.

\section{Conclusion} \label{sec:conclusion}

With visualisations on low-dimensional toy data and extensive experiments covering high-dimensional datasets for anomaly detection, we find that non-bottlenecked AEs (including the \Infbae) can perform reasonably well over the baseline. The major implications of our work are (1) learning the identity function is not as trivial as previously assumed and (2) modellers should not restrict to only bottlenecked architectures since non-bottlenecked architectures can perform better.

In light of the potential of non-bottlenecked AEs, future work should develop more variants. The closed-form solutions of \Infbae can facilitate theoretical work on understanding and proving the conditions for not learning the identity function. Possible directions include understanding the connection between BAEs as predictive density models and kernel density estimation \cite{rosenblatt1956,parzen1962estimation}.

\bibliography{bibliography/xiang, bibliography/xai, bibliography/unc_ood, bibliography/icml, bibliography/bottleneck_ae}

\begin{thebibliography}{10}
\expandafter\ifx\csname url\endcsname\relax
  \def\url#1{\texttt{#1}}\fi
\expandafter\ifx\csname urlprefix\endcsname\relax\def\urlprefix{URL }\fi
\expandafter\ifx\csname href\endcsname\relax
  \def\href#1#2{#2} \def\path#1{#1}\fi

\bibitem{pang2021}
G.~Pang, C.~Shen, L.~Cao, A.~V.~D. Hengel,
  \href{https://doi.org/10.1145/3439950}{Deep learning for anomaly detection: A
  review}, ACM Computing Surveys 54~(2) (mar 2021).
\newblock \href {https://doi.org/10.1145/3439950} {\path{doi:10.1145/3439950}}.
\newline\urlprefix\url{https://doi.org/10.1145/3439950}

\bibitem{ruff2018}
L.~Ruff, R.~Vandermeulen, N.~Goernitz, L.~Deecke, S.~A. Siddiqui, A.~Binder,
  E.~M{\"u}ller, M.~Kloft,
  \href{https://proceedings.mlr.press/v80/ruff18a.html}{Deep one-class
  classification}, in: J.~Dy, A.~Krause (Eds.), Proceedings of the 35th
  International Conference on Machine Learning, Vol.~80 of Proceedings of
  Machine Learning Research, PMLR, 2018, pp. 4393--4402.
\newline\urlprefix\url{https://proceedings.mlr.press/v80/ruff18a.html}

\bibitem{ng2011sparse}
A.~Ng, et~al., Sparse autoencoder, CS294A Lecture notes 72~(2011) (2011) 1--19.

\bibitem{chen2018autoencoder}
Z.~Chen, C.~K. Yeo, B.~S. Lee, C.~T. Lau, Autoencoder-based network anomaly
  detection, in: 2018 Wireless Telecommunications Symposium (WTS), IEEE, 2018,
  pp. 1--5.

\bibitem{chow2020}
J.~Chow, Z.~Su, J.~Wu, P.~Tan, X.~Mao, Y.~Wang,
  \href{https://www.sciencedirect.com/science/article/pii/S1474034620300744}{Anomaly
  detection of defects on concrete structures with the convolutional
  autoencoder}, Advanced Engineering Informatics 45 (2020) 101105.
\newblock \href {https://doi.org/https://doi.org/10.1016/j.aei.2020.101105}
  {\path{doi:https://doi.org/10.1016/j.aei.2020.101105}}.
\newline\urlprefix\url{https://www.sciencedirect.com/science/article/pii/S1474034620300744}

\bibitem{kim2020}
S.~Kim, W.~Jo, T.~Shon,
  \href{https://www.sciencedirect.com/science/article/pii/S1568494619307999}{Apad:
  Autoencoder-based payload anomaly detection for industrial ioe}, Applied Soft
  Computing 88 (2020) 106017.
\newblock \href {https://doi.org/https://doi.org/10.1016/j.asoc.2019.106017}
  {\path{doi:https://doi.org/10.1016/j.asoc.2019.106017}}.
\newline\urlprefix\url{https://www.sciencedirect.com/science/article/pii/S1568494619307999}

\bibitem{mujeeb2019}
A.~Mujeeb, W.~Dai, M.~Erdt, A.~Sourin,
  \href{https://www.sciencedirect.com/science/article/pii/S1474034619301259}{One
  class based feature learning approach for defect detection using deep
  autoencoders}, Advanced Engineering Informatics 42 (2019) 100933.
\newblock \href {https://doi.org/https://doi.org/10.1016/j.aei.2019.100933}
  {\path{doi:https://doi.org/10.1016/j.aei.2019.100933}}.
\newline\urlprefix\url{https://www.sciencedirect.com/science/article/pii/S1474034619301259}

\bibitem{hendrycks2016gaussian}
D.~Hendrycks, K.~Gimpel, Gaussian error linear units (gelus), arXiv preprint
  arXiv:1606.08415 (2016).

\bibitem{scikit-learn}
F.~Pedregosa, G.~Varoquaux, A.~Gramfort, V.~Michel, B.~Thirion, O.~Grisel,
  M.~Blondel, P.~Prettenhofer, R.~Weiss, V.~Dubourg, J.~Vanderplas, A.~Passos,
  D.~Cournapeau, M.~Brucher, M.~Perrot, E.~Duchesnay, Scikit-learn: Machine
  learning in {P}ython, Journal of Machine Learning Research 12 (2011)
  2825--2830.

\bibitem{goodfellow2016deep}
I.~Goodfellow, Y.~Bengio, A.~Courville, Deep Learning, MIT Press, 2016.

\bibitem{chen2014stochastic}
T.~Chen, E.~Fox, C.~Guestrin, Stochastic gradient hamiltonian monte carlo, in:
  International Conference on machine learning, 2014, pp. 1683--1691.

\bibitem{gal2016dropout}
Y.~Gal, Z.~Ghahramani, Dropout as a bayesian approximation: Representing model
  uncertainty in deep learning, in: International Conference on Machine
  Learning, 2016, pp. 1050--1059.

\bibitem{blundell2015weight}
C.~Blundell, J.~Cornebise, K.~Kavukcuoglu, D.~Wierstra, Weight uncertainty in
  neural network, in: International Conference on Machine Learning, PMLR, 2015,
  pp. 1613--1622.

\bibitem{pearce2020uncertainty}
T.~Pearce, F.~Leibfried, A.~Brintrup, Uncertainty in neural networks:
  Approximately bayesian ensembling, in: International conference on artificial
  intelligence and statistics, PMLR, 2020, pp. 234--244.

\bibitem{kingma2013auto}
D.~P. Kingma, M.~Welling, Auto-encoding variational bayes, arXiv preprint
  arXiv:1312.6114 (2013).

\bibitem{unet2015}
O.~Ronneberger, P.~Fischer, T.~Brox, U-net: Convolutional networks for
  biomedical image segmentation, in: N.~Navab, J.~Hornegger, W.~M. Wells, A.~F.
  Frangi (Eds.), Medical Image Computing and Computer-Assisted Intervention --
  MICCAI 2015, Springer International Publishing, Cham, 2015, pp. 234--241.

\bibitem{li2017visualizing}
H.~Li, Z.~Xu, G.~Taylor, C.~Studer, T.~Goldstein, Visualizing the loss
  landscape of neural nets, arXiv preprint arXiv:1712.09913 (2017).

\bibitem{he2016deep}
K.~He, X.~Zhang, S.~Ren, J.~Sun, Deep residual learning for image recognition,
  in: Proceedings of the IEEE conference on computer vision and pattern
  recognition, 2016, pp. 770--778.

\bibitem{baur2020bayesian}
C.~Baur, B.~Wiestler, S.~Albarqouni, N.~Navab, Bayesian skip-autoencoders for
  unsupervised hyperintense anomaly detection in high resolution brain mri, in:
  2020 IEEE 17th International Symposium on Biomedical Imaging (ISBI), IEEE,
  2020, pp. 1905--1909.

\bibitem{collin2021}
A.-S. Collin, C.~De~Vleeschouwer, Improved anomaly detection by training an
  autoencoder with skip connections on images corrupted with stain-shaped
  noise, in: 2020 25th International Conference on Pattern Recognition (ICPR),
  2021, pp. 7915--7922.
\newblock \href {https://doi.org/10.1109/ICPR48806.2021.9412842}
  {\path{doi:10.1109/ICPR48806.2021.9412842}}.

\bibitem{kim2021}
J.~Kim, J.~Ko, H.~Choi, H.~Kim,
  \href{https://www.mdpi.com/1424-8220/21/15/4968}{Printed circuit board defect
  detection using deep learning via a skip-connected convolutional
  autoencoder}, Sensors 21~(15) (2021).
\newblock \href {https://doi.org/10.3390/s21154968}
  {\path{doi:10.3390/s21154968}}.
\newline\urlprefix\url{https://www.mdpi.com/1424-8220/21/15/4968}

\bibitem{neal1996priors}
R.~M. Neal, Priors for infinite networks, in: Bayesian Learning for Neural
  Networks, Springer, 1996, pp. 29--53.

\bibitem{novak2018bayesian}
R.~Novak, L.~Xiao, J.~Lee, Y.~Bahri, G.~Yang, J.~Hron, D.~A. Abolafia,
  J.~Pennington, J.~Sohl-Dickstein, Bayesian deep convolutional networks with
  many channels are gaussian processes, arXiv preprint arXiv:1810.05148 (2018).

\bibitem{yang2019nngp}
G.~Yang, Wide feedforward or recurrent neural networks of any architecture are
  gaussian processes, in: H.~Wallach, H.~Larochelle, A.~Beygelzimer,
  F.~d\textquotesingle Alch\'{e}-Buc, E.~Fox, R.~Garnett (Eds.), Advances in
  Neural Information Processing Systems, Vol.~32, Curran Associates, Inc.,
  2019.

\bibitem{hron2020infinite}
J.~Hron, Y.~Bahri, J.~Sohl-Dickstein, R.~Novak, Infinite attention: Nngp and
  ntk for deep attention networks, in: International Conference on Machine
  Learning, PMLR, 2020, pp. 4376--4386.

\bibitem{chen2006probability}
T.~Chen, J.~Morris, E.~Martin, Probability density estimation via an infinite
  gaussian mixture model: application to statistical process monitoring,
  Journal of the Royal Statistical Society: Series C (Applied Statistics)
  55~(5) (2006) 699--715.

\bibitem{sriperumbudur2017density}
B.~Sriperumbudur, K.~Fukumizu, A.~Gretton, A.~Hyv{\"a}rinen, R.~Kumar, Density
  estimation in infinite dimensional exponential families, Journal of Machine
  Learning Research 18 (2017).

\bibitem{klambauer2017self}
G.~Klambauer, T.~Unterthiner, A.~Mayr, S.~Hochreiter, Self-normalizing neural
  networks, in: Proceedings of the 31st international conference on neural
  information processing systems, 2017, pp. 972--981.

\bibitem{aitchison2020bigger}
L.~Aitchison, Why bigger is not always better: on finite and infinite neural
  networks, in: International Conference on Machine Learning, PMLR, 2020, pp.
  156--164.

\bibitem{lee2020finite}
J.~Lee, S.~Schoenholz, J.~Pennington, B.~Adlam, L.~Xiao, R.~Novak,
  J.~Sohl-Dickstein, Finite versus infinite neural networks: an empirical
  study, Advances in Neural Information Processing Systems 33 (2020).

\bibitem{kingma2014adam}
D.~P. Kingma, J.~Ba, Adam: A method for stochastic optimization, arXiv preprint
  arXiv:1412.6980 (2014).

\bibitem{xiao2017fashion}
H.~Xiao, K.~Rasul, R.~Vollgraf, Fashion-mnist: a novel image dataset for
  benchmarking machine learning algorithms, arXiv preprint arXiv:1708.07747
  (2017).

\bibitem{deng2012mnist}
L.~Deng, The mnist database of handwritten digit images for machine learning
  research, IEEE Signal Processing Magazine 29~(6) (2012) 141--142.

\bibitem{krizhevsky2009learning}
A.~Krizhevsky, G.~Hinton, et~al., Learning multiple layers of features from
  tiny images (2009).

\bibitem{netzer2011}
Y.~Netzer, T.~Wang, A.~Coates, A.~Bissacco, B.~Wu, A.~Y. Ng, Reading digits in
  natural images with unsupervised feature learning, in: NIPS Workshop on Deep
  Learning and Unsupervised Feature Learning 2011, 2011.

\bibitem{rayana2016}
S.~Rayana, \href{http://odds.cs.stonybrook.edu}{{ODDS} library} (2016).
\newline\urlprefix\url{http://odds.cs.stonybrook.edu}

\bibitem{tizian2018}
T.~Schneider, S.~Klein, M.~Bastuck,
  \href{https://doi.org/10.5281/zenodo.1323611}{{Condition monitoring of
  hydraulic systems Data Set at ZeMA}} (Apr. 2018).
\newblock \href {https://doi.org/10.5281/zenodo.1323611}
  {\path{doi:10.5281/zenodo.1323611}}.
\newline\urlprefix\url{https://doi.org/10.5281/zenodo.1323611}

\bibitem{tachtatzis2019}
C.~Tachtatzis, G.~Gourlay, I.~Andonovic, O.~Panni,
  \href{https://doi.org/10.5281/zenodo.3405265}{Sensor data set radial forging
  at afrc testbed v2} (Sep. 2019).
\newblock \href {https://doi.org/10.5281/zenodo.3405265}
  {\path{doi:10.5281/zenodo.3405265}}.
\newline\urlprefix\url{https://doi.org/10.5281/zenodo.3405265}

\bibitem{tukey1977exploratory}
J.~W. Tukey, et~al., Exploratory data analysis, Vol.~2, Reading, Mass., 1977.

\bibitem{smith2017cyclical}
L.~N. Smith, Cyclical learning rates for training neural networks, in: 2017
  IEEE Winter Conference on Applications of Computer Vision (WACV), IEEE, 2017,
  pp. 464--472.

\bibitem{maas2013rectifier}
A.~L. Maas, A.~Y. Hannun, A.~Y. Ng, Rectifier nonlinearities improve neural
  network acoustic models, in: ICML Workshop on Deep Learning for Audio, Speech
  and Language Processing, 2013.

\bibitem{ba2016layer}
J.~L. Ba, J.~R. Kiros, G.~E. Hinton, Layer normalization, arXiv preprint
  arXiv:1607.06450 (2016).

\bibitem{neuraltangents2020}
R.~Novak, L.~Xiao, J.~Hron, J.~Lee, A.~A. Alemi, J.~Sohl-Dickstein, S.~S.
  Schoenholz, \href{https://github.com/google/neural-tangents}{Neural tangents:
  Fast and easy infinite neural networks in python}, in: International
  Conference on Learning Representations, 2020.
\newline\urlprefix\url{https://github.com/google/neural-tangents}

\bibitem{melo2013}
F.~Melo, Area under the ROC Curve, Springer New York, New York, NY, 2013, pp.
  38--39.
\newblock \href {https://doi.org/10.1007/978-1-4419-9863-7_209}
  {\path{doi:10.1007/978-1-4419-9863-7_209}}.

\bibitem{nalisnick2018do}
E.~Nalisnick, A.~Matsukawa, Y.~W. Teh, D.~Gorur, B.~Lakshminarayanan, Do deep
  generative models know what they don't know?, in: International Conference on
  Learning Representations, 2019.

\bibitem{choi2018waic}
H.~Choi, E.~Jang, A.~A. Alemi, Waic, but why? generative ensembles for robust
  anomaly detection, arXiv preprint arXiv:1810.01392 (2018).

\bibitem{hinne2020bma}
M.~Hinne, Q.~F. Gronau, D.~van~den Bergh, E.-J. Wagenmakers,
  \href{https://doi.org/10.1177/2515245919898657}{A conceptual introduction to
  bayesian model averaging}, Advances in Methods and Practices in Psychological
  Science 3~(2) (2020) 200--215.
\newblock \href {https://doi.org/10.1177/2515245919898657}
  {\path{doi:10.1177/2515245919898657}}.
\newline\urlprefix\url{https://doi.org/10.1177/2515245919898657}

\bibitem{zhao2018}
G.~Zhao, J.~Liu, J.~Jiang, H.~Guan, J.-R. Wen, Skip-connected deep
  convolutional autoencoder for restoration of document images, in: 2018 24th
  International Conference on Pattern Recognition (ICPR), 2018, pp. 2935--2940.
\newblock \href {https://doi.org/10.1109/ICPR.2018.8546199}
  {\path{doi:10.1109/ICPR.2018.8546199}}.

\bibitem{mao2016}
X.~Mao, C.~Shen, Y.-B. Yang,
  \href{https://proceedings.neurips.cc/paper/2016/file/0ed9422357395a0d4879191c66f4faa2-Paper.pdf}{Image
  restoration using very deep convolutional encoder-decoder networks with
  symmetric skip connections}, in: D.~Lee, M.~Sugiyama, U.~Luxburg, I.~Guyon,
  R.~Garnett (Eds.), Advances in Neural Information Processing Systems,
  Vol.~29, Curran Associates, Inc., 2016.
\newline\urlprefix\url{https://proceedings.neurips.cc/paper/2016/file/0ed9422357395a0d4879191c66f4faa2-Paper.pdf}

\bibitem{liu2018}
J.-Y. Liu, Y.-H. Yang, Denoising auto-encoder with recurrent skip connections
  and residual regression for music source separation, in: 2018 17th IEEE
  International Conference on Machine Learning and Applications (ICMLA), 2018,
  pp. 773--778.
\newblock \href {https://doi.org/10.1109/ICMLA.2018.00123}
  {\path{doi:10.1109/ICMLA.2018.00123}}.

\bibitem{snoek2012nonparametric}
J.~Snoek, R.~Adams, H.~Larochelle, On nonparametric guidance for learning
  autoencoder representations, in: Artificial Intelligence and Statistics,
  PMLR, 2012, pp. 1073--1080.

\bibitem{nguyen2021benefits}
T.~V. Nguyen, R.~K. Wong, C.~Hegde, Benefits of jointly training autoencoders:
  An improved neural tangent kernel analysis, IEEE Transactions on Information
  Theory (2021).

\bibitem{radhakrishnan2018memorization}
A.~Radhakrishnan, K.~Yang, M.~Belkin, C.~Uhler, Memorization in
  overparameterized autoencoders, arXiv preprint arXiv:1810.10333 (2018).

\bibitem{zhang2019identity}
C.~Zhang, S.~Bengio, M.~Hardt, M.~C. Mozer, Y.~Singer, Identity crisis:
  Memorization and generalization under extreme overparameterization, arXiv
  preprint arXiv:1902.04698 (2019).

\bibitem{rosenblatt1956}
M.~Rosenblatt, \href{https://doi.org/10.1214/aoms/1177728190}{{Remarks on Some
  Nonparametric Estimates of a Density Function}}, The Annals of Mathematical
  Statistics 27~(3) (1956) 832 -- 837.
\newblock \href {https://doi.org/10.1214/aoms/1177728190}
  {\path{doi:10.1214/aoms/1177728190}}.
\newline\urlprefix\url{https://doi.org/10.1214/aoms/1177728190}

\bibitem{parzen1962estimation}
E.~Parzen, On estimation of a probability density function and mode, The annals
  of mathematical statistics 33~(3) (1962) 1065--1076.

\end{thebibliography}



\end{document}